\documentclass[conference]{IEEEtran}
\IEEEoverridecommandlockouts
\usepackage{cite}
\usepackage{amsmath,amssymb,amsfonts}
\usepackage{algorithmic,algorithm}
\usepackage{graphicx}
\usepackage{textcomp}
\usepackage{xcolor}
\usepackage{url}

\def\BibTeX{{\rm B\kern-.05em{\sc i\kern-.025em b}\kern-.08em
    T\kern-.1667em\lower.7ex\hbox{E}\kern-.125emX}}

\DeclareMathOperator*{\argmin}{arg\,min}

\begin{document}

\title{Accuracy of TextFooler black box adversarial attacks on 01 loss sign activation neural network ensemble\\
}

\author{\IEEEauthorblockN{1\textsuperscript{st} Yunzhe Xue}
\IEEEauthorblockA{\textit{Department of Data Science} \\
\textit{New Jersey Institute of Technology}\\
Newark, NJ, USA \\
yx277@njit.edu}
\and
\IEEEauthorblockN{2\textsuperscript{nd} Usman Roshan}
\IEEEauthorblockA{\textit{Department of Data Science} \\
\textit{New Jersey Institute of Technology}\\
Newark, NJ, USA \\
usman@njit.edu}
}

\maketitle

\begin{abstract}
Recent work has shown the defense of 01 loss sign activation neural networks against image classification adversarial attacks. A public challenge to attack the models on CIFAR10 dataset remains undefeated. We ask the following question in this study: are 01 loss sign activation neural networks hard to deceive with a popular black box text adversarial attack program called TextFooler? We study this question on four popular text classification datasets: IMDB reviews, Yelp reviews, MR sentiment classification, and AG news classification. We find that our 01 loss sign activation network is much harder to attack with TextFooler compared to sigmoid activation cross entropy and binary neural networks. We also study a 01 loss sign activation convolutional neural network with a novel global pooling step specific to sign activation networks. With this new variation we see a significant gain in adversarial accuracy rendering TextFooler practically useless against it. We make our code freely available at \url{https://github.com/zero-one-loss/wordcnn01} and \url{https://github.com/xyzacademic/mlp01example}. Our work here suggests that 01 loss sign activation networks could  be further developed to create fool proof models against text adversarial attacks.
\end{abstract}


\section{Introduction}
Adversarial attacks remain a security vulnerability in neural networks today since they were first discovered \cite{szegedy2013intriguing}. A recent paper evaluating the operational feasibility of adversarial attacks in military defense \cite{RR-A866-1} found that patch attacks posed a minimal danger in practice but both white and black-box attacks can be significantly more effective - a white box attack reduces the target model's accuracy by 65\% and a black-box attack reduces it by 55\% to 63\% depending upon knowledge and access to training data and target model architecture. Various defenses have been developed and broken \cite{athalye2018obfuscated}. Adversarial training \cite{kurakin2016adversarial}, which is to train the model with clean and adversarial data, remains the most effective solution to date but it is computationally expensive and lowers clean test accuracy \cite{raghunathan2019adversarial,clarysse2022adversarial}.

Recent work has shown that 01 loss sign activation networks are hard to attack in image classification datasets \cite{xie2019,xue2020transferability,xue2020towards,yang2020accurate,yang2022defense,xue2023}. In this paper we investigate its robustness against text black box adversrial attacks.

The TextFooler \cite{jin2020bert} method is designed to find syntactically and semantically similar adversarial documents by replacing important words with similar ones until the document is misclassified. We attack models with TextFooler on four document classification datasets: Internet Movie Database (25K train, 25K test, mean words per document=215) and Yelp (560K train, 38K test, mean words per document=152) positive and negative reviews (IMDB and Yelp), sentence classification of positive and negative sentiments (9K train, 1K test, mean words per document=20, denoted as MR), and sentence-level classification of news items in World and Sports categories (120K train, 7.6K test, mean words per document=43, denoted as AG) \cite{jin2020bert}. 

\section{Methods}

\subsection{Gradient-free stochastic coordinate decent for sign activation 01 loss network}

Suppose we are given binary class data $x_i \in R^d$ and $y_i \in \{-1,+1\}$ for $i=0...n-1$. Consider the objective function of a single hidden layer neural network with sign activation and 01 loss given below. We employ a stochastic coordinate descent algorithm shown in Algorithm~\ref{mlp01} (similar to recent work \cite{xue2020towards,xue2020transferability,xie2019}) to minimize this objective. 

\begin{equation}
\frac{1}{2n}\argmin_{W, W_0, w,w_0} \sum_i (1-sign(y_i(w^T(sign(W^Tx_i+W_0))+w_0)))
\label{obj2}
\end{equation}

\begin{algorithm}[!h]
\caption{SCD01: Stochastic coordinate descent for sign activation 01 loss single hidden layer network} 
\label{mlp01}
\textbf{Procedure: }
\begin{algorithmic}
\STATE 1. Initialize all network weights $W,w$ to random values from the Normal distribution $N(0,1)$.
\STATE 2. Set network thresholds $W_0$ to the median projection value on their corresponding weight vectors and $w_0$ to the projection value that minimizes our network objective.
\WHILE {$i < epochs$} 
	\STATE 1. Randomly sample 75\% of data equally from each class.
	\STATE 2. Perform coordinate descent separately first on the final node $w$ and then a randomly selected hidden node $u$ (a random column from the hidden layer weight matrix $W$)
	\STATE 3. Suppose we are performing coordinate descent on node $w$. We select a random set of features (coordinates) from $w$ called $F$. For each feature $w_i \in F$ we add/subtract a learning rate $\eta$ and then determine the $w_0$ that optimizes the loss (done in parallel on a GPU). We consider all possible values of $w_0=\frac{w^Tx_i + w^Tx_{i+1}}{2}$ for $i=0...n-2$ and select the one that minimizes the loss (also performed in parallel on a GPU).
	\STATE 4. After making the update above we evaluate the loss on the full dataset (performed on a GPU for parallel speedups) and accept the change if it improves the loss. 
\ENDWHILE
\end{algorithmic}
\end{algorithm}

We can train sign activation networks with and without binary weights using our SCD training procedure above. In the case of binary weights we don't need a learning rate. We apply GPU parallelism to simultaneously update features and other heuristics to speed up runtimes.

\subsection{Implementation, test accuracy, and runtime}
We implement our training procedure in Python, numpy, and Pytorch \cite{pytorch}.
and make our code freely available from \url{https://github.com/zero-one-loss/wordcnn01} and \url{https://github.com/xyzacademic/mlp01example}. Since sign activation is non-convex and our training starts from a different random initialization we run it 8 times and output the majority vote. 

To illustrate our real runtimes and clean test accuracies we compare our model with a single hidden layer of 20 nodes to the equivalent network with sigmoid activation and logistic loss (denoted as MLP) and the binary neural network (denoted as BNN) \cite{hubara2016binarized}. We used the MLPClassifier in scikit-learn \cite{scikit} to implement MLP and the Larq library \cite{geiger2020larq} with the \emph{approx} approximation to the sign activation. This has shown to achieve a higher test accuracy than the original straight through estimator (STE) of the sign activation \cite{liu2018bi}. 

We perform a 1000 iterations of SCD01. Our training runtimes are comparable to gradient descent in MLP and BNN and thus practically usable. We can trivially parallelize training an ensemble by doing multiple runs on CPU and GPU cores at the same time. 

\section{Results}
We include in our experiments a WordCNN with sigmoid activation and cross-entropy loss (described below) denoted as CNN and a random forest classifier  \cite{breiman2001random} denoted as RF. In Table~\ref{table1} we see that ensembles of our models give the highest adversarial accuracy on all four datasets and require the greatest number of queries. If a smaller limit was placed on the allowed queries (for example imposed by the system being attacked) we can expect a higher adversarial accuracy for our models. Here we show ensembles of 8 votes for each model - in the random forest we also use 8 trees. 

\begin{table}[htbp]
\caption{Accuracy of clean and TextFooler black-box adversarial examples denoted by Cl and Adv respectively. All models shown here are 8 votes. Also shown are the number of queries made by the attacker denoted as Que. We round accuracies and queries to the nearest integer. For each dataset highest adversarial accuracy shown in bold. \label{text}}
\begin{center}
\begin{tabular}{llll | lll}
 &  \multicolumn{3}{c}{\bf IMDB}  & \multicolumn{3}{c}{\bf Yelp}  \\ 
   &  \bf Cl & \bf Adv & \bf Que & \bf Cl & \bf Adv &  \bf Que \\ \hline \\
SCD01 & 82 & {\bf 51} & 3279 & 85 & {\bf 54} & 1908 \\
CNN & 89.2 & 0 & 524 & 94 & 1.1 & 492  \\
MLP & 85 & 0 & 686 & 87.3 &. 2 & 500  \\
BNN & 83.8 & 21.5 & 2301 & 85 & 32.3 & 1622  \\
RF & 76.7 & 11 & 1823 & 77.7 & 7.5 &  935  \\ \hline
\end{tabular}\\
\begin{tabular}{llll | lll | lll | lll}
 & \multicolumn{3}{c}{\bf MR} & \multicolumn{3}{c}{\bf AG} \\ 
 & \bf Cl & \bf Adv & \bf Que & \bf Cl & \bf Adv & \bf Que \\  \hline \\
SCD01 & 74 & {\bf 14} & 186 & 99 & {\bf 93} & 672 \\
CNN & 78 & 2.8 & 123 & 96.5 & 	49.1 & 258 \\
MLP &  75 & 2.3 & 123 & 99 & 51.4 & 366 \\
BNN &  73.2 & 5.8 & 150 & 99.1 & 75.6 & 564  \\
RF &  68.1 & 2.1 & 115 & 96.7 & 72.2 & 532 \\ \hline
\end{tabular}\label{table1}
\end{center}
\end{table}

\begin{figure*}[h!]
\centerline{\includegraphics[trim={0 0.5in 0 0},clip,width=.9\linewidth]{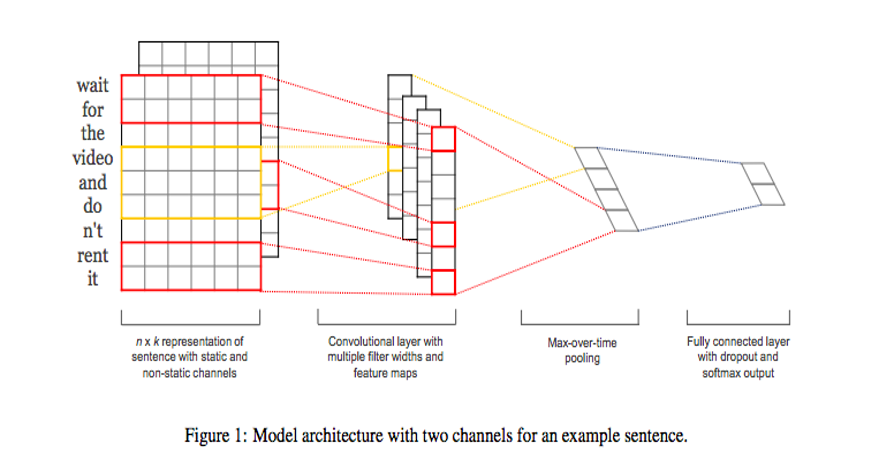}} 
\caption{CNN text classiication model - the model above shows two input channels for a given sentence}
\label{wordcnn}
\end{figure*}

WordCNN stacks word vectors \cite{pennington2014glove} of each word in a document into a matrix to treat it as  2D image \cite{kim2014convolutional} (see Figure~\ref{wordcnn}). In the other models that take feature vectors as inputs we consider the averaged word vector of all words in a document \cite{lilleberg2015support}. For all models we use 200 dimensional Glove word embeddings pre-trained on 6 billion tokens from Wikipedia and Gigawords \cite{pennington2014glove}. This gives a lower clean test accuracy than WordCNN but still above an acceptable level in practice. 

With sign activation and 01 loss the CNN becomes CNN01. We study a new variation of it where instead of global average pooling over +1 and -1 we only sum the 1’s. We call this CNN01-FS. For example if we have two sentences of length 10 and 100 each with 4 keywords, by averaging the 4 words we have weights 40\% and 4\% respectively (lower in the larger sentence), but by summing it’s the same in both. We find this to significantly improve the adversarial accuracy of CNN01. In Table\~ref{table2} we see that CNN01-FS improves several orders of magnitude upon the original one. Even the CNN with sigmoid activation has a slight improvement with the variation but not nearly as large as CNN01.

\begin{table}[h!]
\caption{Accuracy of clean and TextFooler black-box adversarial examples denoted by Cl and Adv respectively. All models shown here are 8 votes. Also shown are the number of queries made by the attacker denoted as Que. We round accuracies and queries to the nearest integer. For each dataset highest adversarial accuracy shown in bold. \label{text}}
\begin{center}
\begin{tabular}{llll | lll | lll | lll}
 &  \multicolumn{3}{c}{\bf IMDB}  & \multicolumn{3}{c}{\bf Yelp} \\ 
   &  \bf Cl & \bf Adv & \bf Que & \bf Cl & \bf Adv &  \bf Que \\  \hline \\
CNN01-FS &  80.7 & {\bf 67.9} & 1288 & 89.6 & {\bf 64.6} & 566 \\ 
CNN01 &  84.9 & 0.2 & 299 & 89.4 & 19.5 & 296  \\ 
CNN-FS & 88.4 & 8.5 & 419 & 92.9 & 3 & 220 \\
CNN & 86.9 & 0.3 & 301 & 92.2 & 10.7 & 252  \\
\end{tabular}
\begin{tabular}{llll | lll | lll | lll}
 & \multicolumn{3}{c}{\bf MR} & \multicolumn{3}{c}{\bf AG} \\ 
 & \bf Cl & \bf Adv & \bf Que & \bf Cl & \bf Adv & \bf Que \\
   \hline \\
CNN01-FS &  76.6 & {\bf 40.9} & 265 & 84.3 & {\bf 66.7} & 484 \\ 
CNN01 &  77.6 & 13.1 & 157 & 85.7 & 5.7 & 207 \\ 
CNN-FS &  79.1 & 14.9 & 170  & 85.8 & 4.3 & 214 \\
CNN & 79.4 & 9.2 & 142 & 86.9 & 3.3 & 208 \\
\end{tabular}
\label{table2}
\end{center}
\end{table}

\section{Discussion}
One reason why our models are hard to deceive is that TextFooler relies upon output probabilities to craft its attack. Our models focus on the sign of the outputs in their loss. We output probabilities TextFooler needs by counting the number of 0 and 1 outputs in the ensemble. However, these probabilities are not useful enough for TextFooler to volley an effective attack.

\section{Conclusion}
Our work here shows that 01 loss sign activation network ensembles are hard to deceive with TextFooler text adversarial attacks. This is consistent with earlier work on image adversarial attacks on the same model. Going forward this may serve as a helpful baseline for robust secure AI models.


\bibliographystyle{IEEEtran}
\bibliography{IEEEabrv,my_bib}  

\end{document}